\documentclass[letterpaper]{article} 
\usepackage{aaai2026}  
\usepackage{times}  
\usepackage{helvet}  
\usepackage{courier}
\usepackage{amsmath}
\usepackage[hyphens]{url}  
\usepackage{graphicx} 
\usepackage{natbib}  
\usepackage{caption} 
\usepackage{booktabs}     
\usepackage{multirow}     
\usepackage{colortbl}      
\usepackage[table]{xcolor} 

\DeclareMathOperator*{\argmax}{arg\,max}
\usepackage[utf8]{inputenc} 
\usepackage{algorithm}
\usepackage{algpseudocode}
\usepackage{amsmath}
\usepackage{amssymb}
\usepackage{xcolor}
\usepackage{newfloat}
\usepackage{listings}

\DeclareCaptionStyle{ruled}{labelfont=normalfont,labelsep=colon,strut=off} 
\floatstyle{ruled}
\newfloat{listing}{tb}{lst}{}
\floatname{listing}{Listing}
\title{Iterative Zoom-In: Temporal Interval Exploration for Long Video Understanding}

\author{
    Chenglin Li \quad Qianglong Chen \quad fengtao \quad Yin Zhang\\
    Zhejiang University\\
    Hangzhou, China\\
    {\tt\small \{lichenglin, chenqianglong, fengtao, zhangyin\}@zju.edu.cn}
}

\usepackage{bibentry}

\begin{document}

\maketitle

\begin{abstract}
Multimodal Large Language Models (MLLMs) have shown strong performance in video understanding tasks. However, they continue to struggle with long-form videos because of an inefficient perception of temporal intervals. Unlike humans, who can dynamically adjust their temporal focus to locate query-relevant moments, current MLLMs often rely on dense, uniform sampling across the video timeline, leading to high memory consumption and a risk of missing crucial information. To address this challenge, we introduce Temporal Search (TS), a training-free framework that enables MLLMs to iteratively explore temporal regions for improved long-video understanding. TS is based on a key observation: the model’s generation confidence across different temporal intervals is highly correlated with prediction accuracy. TS operates through two main iterative stages. First, the MLLM proposes a temporal interval that is that is likely to contain task-relevant information. Then, it samples a fixed number of frames from the interval, regardless of length, and feeds them into the model to produce a refined response and confidence score. TS refines the focus of the model by iteratively shifting attention to more fine-grained temporal intervals, improving its understanding of long videos.
Additionally, keyframe-level descriptions are collected to facilitate cross-interval perception throughout the video. To further improve efficiency, we introduce TS-BFS, a best-first search strategy over a tree. Each node represents a candidate interval and is expanded via two methods: self-driven proposals and uniform partitioning. Nodes are scored based on confidence and self-evaluation, and the most promising one is selected for continued exploration. Extensive experiments on long-form video QA benchmarks demonstrate that TS consistently improves the performance of MLLMs on long videos. Specifically, TS-BFS boosts Qwen-VL-2.5’s accuracy from 51.5\% to 57.9\% on LongVideoBench, and from 48.5\% to 55.1\% on VideoMME.
\end{abstract}


\begin{figure}[t]
\centering
\includegraphics[width=1.0\columnwidth]{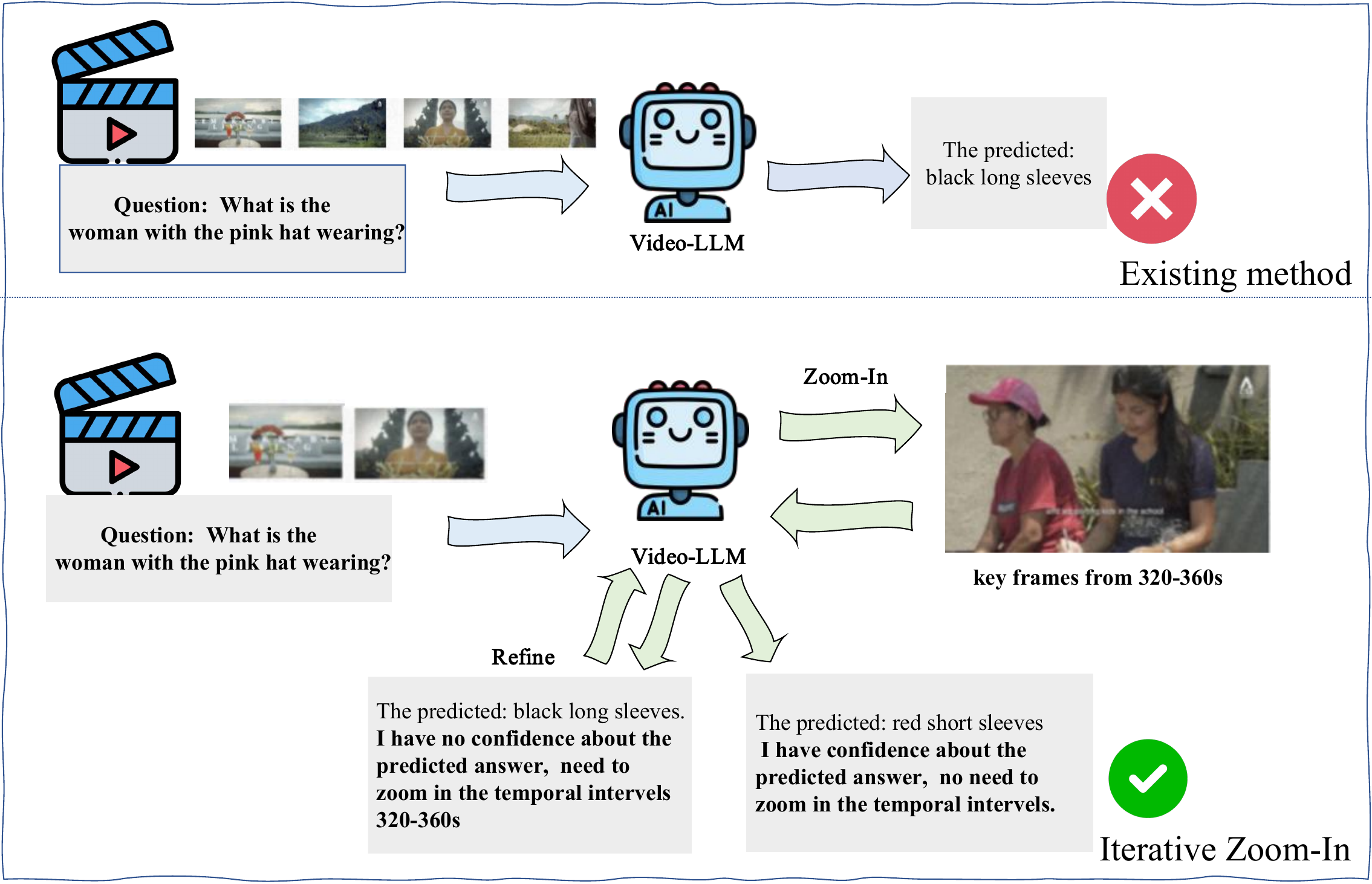} 
\caption{Iterative Zoom-In vs. Single-Pass Inference. Our method iteratively proposes new temporal intervals and samples a fixed number of frames from each. Shorter intervals lead to finer temporal perception, effectively zooming in on relevant intervals for more accurate understanding.}
\label{fig:compare}
\end{figure}

\begin{figure*}[t]
\centering
\includegraphics[width=1.0\textwidth]{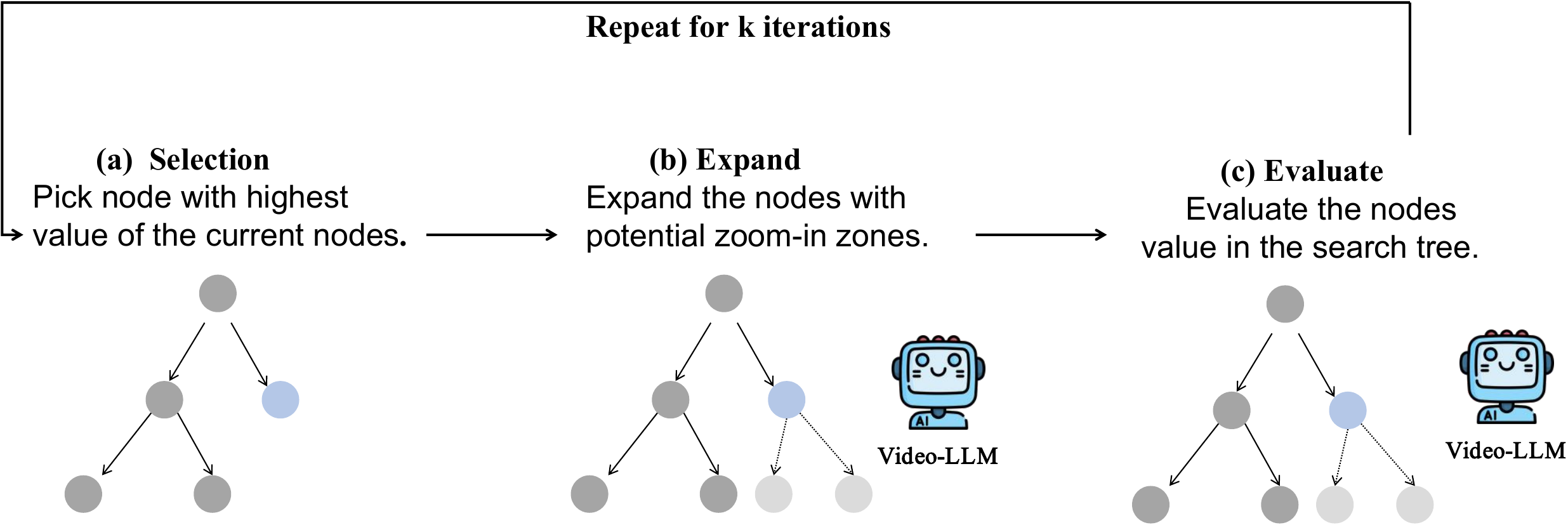} 
\caption{Nodes are selected based on a combined score of the model’s prediction confidence and its self-evaluation of answer correctness. Each node is expanded using two strategies: model-guided proposal and uniform splitting of the current interval. In each iteration, the VideoLLM proposes new temporal intervals, generates responses with confidence scores, evaluates its answers, and produces keyframe-level descriptions incorporated into the input to support global video understanding.}
\label{fig:frame}
\end{figure*}
\section{Introduction}
Multimodal Large Language Models (MLLM)~\cite{li2023videochat,zhang2023video,lin2023video,li2024llava,bai2025qwen2,zhang2024video} have shown strong capabilities in video understanding tasks~\cite{yu2019activitynet,ning2023video,chen2023autoeval,fang2024mmbench,li2024videovista,li2024mvbench}. However, they still face challenges when dealing with long-form videos~\cite{fu2024video,wu2024longvideobench}, largely due to the computational cost associated with processing a large number of frames. A common practice is to apply uniform dense sampling in the video timeline~\cite{tang2025video,nguyen2024video,xiao2025videoqa,huang2024image}. While simple, this approach is memory-intensive and often fails to capture the fine-grained temporal cues essential for long-form understanding~\cite{cheng2025scaling}. In contrast, humans tend to analyse videos more selectively. Rather than inspecting all frames, they intuitively focus on salient temporal regions~\cite{zacks2007event}. When visual information is not clear, humans tend to iteratively focus on fine-grained temporal regions until a satisfactory answer is found~\cite{frederick2005cognitive,paternoster2009rational}. Inspired by this behaviour, we develop a temporal search strategy that iteratively explores diverse regions of the video and refines predictions until a confident conclusion is reached. In particular, we observe that the confidence in the model generation over a temporal interval is highly correlated with its prediction accuracy. Building on this insight, we propose Temporal Search (TS), a search-based framework for efficient and accurate long-form video understanding as shown in Figure~\ref{fig:compare}. 
  
TS iteratively proposes temporal intervals that are likely to contain task-relevant content and uniformly samples a fixed number of frames to feed into the MLLM regardless of the length of the interval. As intervals become shorter, the input captures finer-grained temporal details, enabling the model to zoom in on critical temporal regions iteratively. To further improve the iterative process, we introduce TS-BFS, a best-first search strategy over a temporal interval tree (TS-BFS).
Each node represents a temporal segment defined by its start and end frames. The search involves three key operations: selection, expansion, and evaluation, as shown in Figure~\ref{fig:frame}. The root node is initialised with frames sampled from the entire video. At each step, the most promising node is selected based on a combined score of confidence in the model and self-evaluation. Expansion is conducted by either the model proposing new intervals or by uniformly splitting the current interval. During evaluation, the model estimates the likelihood that its response is correct, based on the available information. In addition, it generates keyframe-level descriptions relevant to the query, which are stored in a global memory for cross-segment awareness. During the search, a priority queue ranks the candidate nodes, and the one with the highest score is selected for expansion. Either reaching the maximum number of iterations or achieving a sufficiently high confidence score. Unlike existing methods that process many uniformly sampled frames in a single pass, TS adopts an efficient iterative strategy that uses only a minimal number of frames per step. We evaluate TS on multiple long-form video understanding benchmarks, including LongVideoBench~\cite{wu2024longvideobench} and VideoMME~\cite{fu2024video}, achieving consistent accuracy gains over single-pass video perception with models such as Qwen-VL-2.5~\cite{bai2025qwen2}, LLaVA-OV~\cite{li2024llava}, and LLaVA-Video~\cite{zhang2024videoinstructiontuningsynthetic}. Specifically, TS improves Qwen-VL-2.5’s accuracy from 51.5\% to 57.9\% on LongVideoBench, and from 48.5\% to 55.1\% on VideoMME.


\section{Related Works}
\subsection{Large Video-Language Models}
Most Video-LLMs extend image-based multimodal models~\cite{liu2023visual} by incorporating multiple frames to capture temporal dynamics~\cite{li2023videochat,zhang2023video,lin2023video,li2024llava}. These models typically consist of a frame-wise visual encoder, followed by a projection module (e.g., MLP or Q-Former) to align visual features with the language model~\cite{nievideo}. This pipeline has proven effective on short video clips. However, as the number of frames increases, the memory and computational cost of frame-by-frame encoding grows rapidly~\cite{xu2024pllava}. Additionally, dense and uniform frame sampling results in excessive redundancy, hindering video understanding performance.
To alleviate this, recent studies address the challenge of limited input frame numbers by extending the context length of models~\cite{chen2024longvila} or explore frame-level and token-level compression techniques such as adaptive pooling and token pruning~\cite{liu2024ppllava,wu2024videollm,wang2024dynamic,alvar2025divprune}, thus significantly expanding the effective input frame number. Although these approaches reduce resource usage, they often degrade performance by discarding critical temporal details. In contrast, we propose Temporal Search (TS), a training-free framework that formulates temporal perception as a search problem. Instead of processing the entire video at once, TS performs inference at the segment level, iteratively selecting and reasoning over diverse temporal regions. Each iteration operates on a small, memory-efficient set of frames, avoiding the need for dense input and improving both efficiency and accuracy.

\subsection{Tree Search with Large Language Models}
Tree search algorithms have long been effective for navigating structured decision spaces, with strategies such as depth-first, breadth-first~\cite{korf1985depth}, best-first~\cite{dechter1985generalized} widely used in planning and game-playing tasks. Recent research has explored their integration with Large Language Models (LLMs) to enhance reasoning and generation. By iteratively refining outputs through search, LLMs can better explore alternative answers and self-correct over time. This paradigm often referred to as self-refinement or tree-of-thoughts~\cite{yao2023tree}, has been shown to improve performance on complex question answering and decision-making tasks. In the multimodal domain, \cite{li2025dyfo,shen2024zoomeye} apply visual search techniques for image understanding tasks. Similarly, tree search methods have also been applied to video understanding.
 Existing studies like VideoTree~\cite{wang2024videotree} and VideoAgent~\cite{wang2024videoagent} combine LLMs with external tools such as captioners or retrievers modules to process long videos. However, these pipelines often limit LLMs to controller or post-processing roles, rarely enabling direct reasoning over a video’s temporal structure. In contrast, we treat temporal understanding as an active, internal search process.
We propose a lightweight, training-free tree search framework that empowers VideoLLMs to direct their temporal search. This approach extends recent LLM-based self-refine techniques~\cite{yao2023tree,hao2023reasoning,zhang2024accessing} to long-form video reasoning with minimal external dependencies.


\begin{algorithm}[t]
\caption{Vanilla Temporal Search (TS)}
\label{alg:ts}
\small
\begin{algorithmic}[1]
\State \textbf{Input:} Video $V = \{f_1, \dots, f_T\}$; query $Q$; steps $k$; thresholds $c_1$, $c_2$; model $p_\theta$
\State \textbf{Initialization:} $I_0 = [0, T]$, keyframes $\mathcal{K} \gets \emptyset$
\State $(\hat{y},\, \gamma) \gets p_\theta(I_0, Q,\, \mathcal{K})$
\If{$\gamma > c_1$}
    \State \Return $\hat{y}$
\EndIf

\For{$i = 1$ to $k$}
    \State $I_i \gets p_\theta(I_{i-1}, Q,\, \mathcal{K},\, \mathit{Prompt}_{\text{expand}})$
    \State $(\hat{y},\, \gamma) \gets p_\theta(I_i, Q,\, \mathcal{K})$
    \If{$\gamma > c_1$}
        \State \Return $\hat{y}$
    \ElsIf{$\gamma > c_2$}
        \State $d \gets p_\theta(I_i, Q,\, \mathit{Prompt}_{\text{keyinfo}})$
        \State $\mathcal{K} \gets \mathcal{K} \cup \{d\}$
    \EndIf
\EndFor

\State \Return $\hat{y}$
\end{algorithmic}
\end{algorithm}

\begin{algorithm}[t]
\caption{Temporal Search with Best-First Tree-Based Search (TS-BFS)}
\label{alg:ts-bfs}
\small
\begin{algorithmic}[1]
\State \textbf{Input:} Video $V = \{f_1, \dots, f_T\}$; query $Q$; steps $k$; thresholds $c_1$, $c_2$; intervals per expand $n$; frames per call $n_f$; model $p_\theta$; 

\State \textbf{Initialization:} $\mathcal{Q} \gets \{\text{Node}(0, T)\}$, $\mathcal{K} \gets \emptyset$
\State Sample $\{f_i\} \sim \text{Uniform}(0, T)$
\State $(\hat{y}, \gamma) \gets p_\theta(\{f_i\}, q, \mathcal{K})$
\State $s \gets p_\theta(\{f_i\}, \mathit{prompt}_{\text{evaluate}}, Q, \hat{y}, \mathcal{K})$
\State $\text{Root.update}(\hat{y}, s + \gamma)$
\If{$\gamma > c_1$}
    \State \Return $\hat{y}$
\EndIf

\For{$t = 1$ to $k$}
    \State $(s^*, e^*) \gets \mathcal{Q}.\text{pop}()$
    \State Sample $\{f_i\} \sim \text{Uniform}(s^*, e^*)$
    \State $\mathcal{C} \gets \mathit{interval}_{\text{heur}} \cup \mathit{interval}_{\text{uniform}}$
    \State \hskip1em $\mathit{interval}_{\text{heur}} \gets p_\theta(\{f_i\}, \mathit{Prompt}_{\text{expand}}, Q, \mathcal{K})$
    \State \hskip1em $\mathit{interval}_{\text{uniform}} \gets \text{UniformSplit}(s^*, e^*, n)$
    \State $\mathcal{S} \gets \text{Sample}(\mathcal{C}, n)$

    \For{$(s', e') \in \mathcal{S}$}
        \State Sample $\{f'_i\} \sim \text{Uniform}(s', e')$
        \State $(\hat{y}', \gamma') \gets p_\theta(\{f'_i\}, q, \mathcal{K})$
        \State $e' \gets p_\theta(\{f'_i\}, \mathit{Prompt}_{\text{evaluate}}, Q, \hat{y}', \mathcal{K})$
        \State $\mathcal{Q}.\text{push}(\text{Node}(s', e', \hat{y}',\, e' + \gamma'))$
        \If{$\gamma' > c_2$}
            \State $d \gets p_\theta(\{f'_i\}, \mathit{Prompt}_{\text{keyinfo}}, q, \mathcal{K})$
            \State $\mathcal{K} \gets \mathcal{K} \cup \{d\}$
       \ElsIf{$\gamma' > c_1$}
            \State \Return $\hat{y}'$
        \EndIf
    \EndFor
\EndFor

\State $\hat{n} \gets \argmax_{\mathit{node} \in \mathcal{Q}} \mathit{node}.\text{value}$
\State \Return $\hat{n}.\text{pred}$
\end{algorithmic}
\end{algorithm}

\section{Method}


\subsection{Problem Setting}
In our setting, a large video-language model (VideoLLM), denoted as $p_{\theta}$ takes as input a long-form video $V = \{f_1, f_2, \dots, f_T\}$, represented as a sequence of frames, and a natural language query $Q$. The model generates a textual response $A = [a_1, \dots, a_m]$ in an autoregressive manner, where each token $a_t$ is sampled from the conditional distribution:
\begin{equation}
a_t \sim p_{\theta}(a_t \mid V, Q, a_{<t}) \propto \exp\left(\operatorname{logit}_\theta(a_t \mid V, Q, a_{<t})\right),
\end{equation}
where $\operatorname{logit}_\theta$ denotes the unnormalized logit (i.e., pre-softmax score) assigned to token $a_t$ given the context. Prior approaches often densely and uniformly sample frames from the entire video $V$, leading to memory-intensive computation that is impractical due to GPU constraints and the sparsity of relevant content over time. To this end, we reformulate long-video understanding as a temporal search problem, where the objective is to locate key temporal intervals in $I$, ultimately enabling accurate answer generation. Let $I = \{f_i, \dots, f_j\}$ denote a candidate interval. Given $I$ and $Q$, the model generates a response $\hat{A} = [\hat{a}_1, \dots, \hat{a}_m]$ along with a confidence score that reflects the model’s certainty in the generated answer. This score is computed as the average log-probability of the generated tokens:

\begin{equation}
\text{Conf}(I, Q) = \frac{1}{m} \sum_{t=1}^{m} \log p_{\theta}(\hat{a}_t \mid I, Q, \hat{a}_{<t})
\end{equation}

\begin{figure}[t]
\centering
\includegraphics[width=0.42\textwidth]{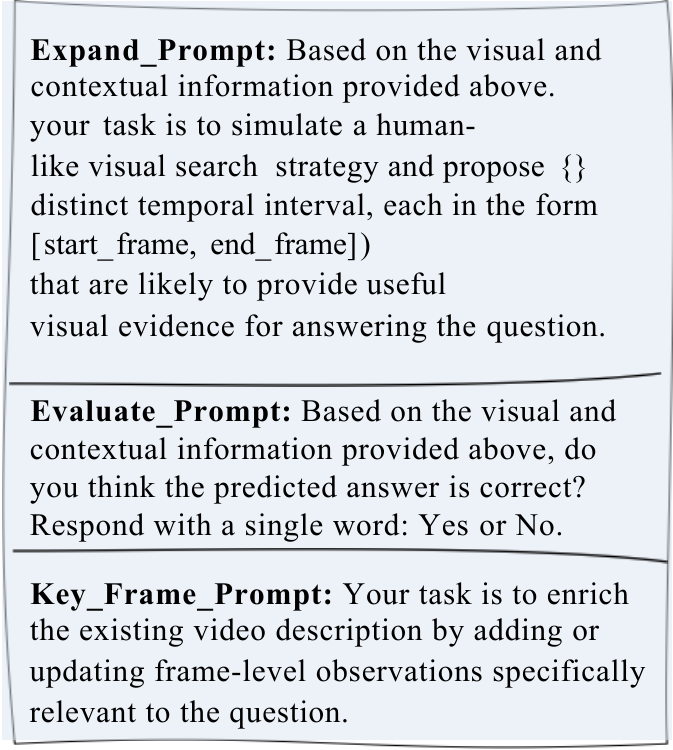} 
\caption{Prompts  for Tree Search.}
\label{fig:prompts}
\end{figure}

 We empirically observe that higher generation confidence over a temporal interval correlates with greater prediction accuracy, making it a reliable signal for guiding segment-level search in long videos. We also design three prompts through prompt engineering, \texttt{Expand\_Prompt}, \texttt{Evaluate\_Prompt}, and \texttt{Key\_Frame\_Prompt} to guide node expansion, evaluation, and keyframe description generation as shown in Figure~\ref{fig:prompts}.
Keyframe descriptions will be integrated into the input prompt, enabling the VideoLLM to perceive the video globally.

\subsection{Temporal Search Framework}
To efficiently explore and refine relevant temporal regions in long videos, we propose Temporal Search (TS), a training-free framework that progressively focuses attention on informative segments. TS integrates two zoom-in strategies: 
\begin{enumerate}
    \item \textbf{Sequential Temporal Search (TS):} A lightweight, step-wise approach that proposes one new temporal interval at each step and directly refines the response within it.
    \item \textbf{Best-First Tree-Based Search (TS-BFS):} A tree search framework that proposes multiple candidate intervals per step, evaluates them, and selects the most promising one for response refinement and further exploration.
\end{enumerate}
\paragraph{Sequential Temporal Search.}
The vanilla TS strategy starts by treating the entire video as the initial interval \( I_0 = [0, T] \), where \( T \) is the total number of frames. The model \( p_\theta \) generate an initial prediction \( \hat{y}_0 \) and a corresponding confidence score \( \gamma_0 = \operatorname{Conf}(I_0, Q) \), given a query \( Q \). If \( \gamma_0 \leq c_1 \), the search is refined by proposing a new interval \( I_1 = [s_1, e_1] \). The model then uniformly samples \( n \) frames from \( I_1 \), performs inference again, and outputs an updated prediction \( \hat{y}_1 \) along with a new confidence score \( \gamma_1 = \operatorname{Conf}(I_1, Q) \). This refinement process is repeated iteratively: at each step \( i \), a new interval \( I_i \) is proposed based on prior predictions, and the model generates a corresponding \( \hat{y}_i \) and \( \gamma_i \).  It also collects keyframe-level descriptions to enhance the overall comprehension. The search terminates when either \( \gamma_i > c_1 \) or a maximum number of iterations \( k \) is reached as shown in Algorithm~\ref{alg:ts}. 
\paragraph{Best-First Tree-Based Search (TS-BFS).}
To further enhance the initial zoom-in process, we introduce a best-first tree-based search strategy (TS-BFS) as shown in Algorithm~\ref{alg:ts-bfs}. Each node in TS-BFS represents a temporal interval defined by its start and end frame indices \( [s, e] \), along with a predicted response and associated confidence score. A fixed number of frames is uniformly sampled from each segment, regardless of its duration, which ensures that longer intervals provide a coarse summary, while shorter intervals capture finer temporal details. 
More promising intervals are prioritised for zoom-in, with the top-score node selected for expansion. The key steps of this process are detailed below.
\paragraph{Node Expansion.}
To progressively focus on informative content, a selected node $[s, e]$ is expanded into child intervals using two strategies:
\begin{itemize}
    \item \textbf{Heuristic Proposal:} The VideoLLM proposes $n$ sub-intervals based on its internal assessment of where relevant information may reside with \texttt{Expand\_Prompt}.
    \item \textbf{Uniform Partitioning:} The interval is evenly split into $n$ sub-segments.
\end{itemize}
Each resulting sub-interval becomes a new node and is evaluated in the same manner.

\paragraph{Node Evaluation.} 
For each node, a fixed number of frames are uniformly sampled and input into the VideoLLM, which generates a predicted answer $y$ and a confidence score $c$. Empirically, we observe that higher confidence scores are strongly correlated with greater answer accuracy. Thus, $c$ is used as a value metric for prioritizing nodes. The confidence score is derived from the model’s token-level probabilities. For a predicted answer $y = [y_1, \dots, y_m]$, the confidence score is computed as $\operatorname{Conf}(S, Q)$, where $S$ denotes the temporal segment associated with the node, and $Q$ is the query.
In addition, \texttt{Evaluate\_Prompt} is used to instruct the VideoLLM to assess the quality of the generated answer. Specifically, the model is prompted to produce a binary judgment (e.g., \texttt{yes} or \texttt{no}), and the probability assigned to the \texttt{yes} response is taken as the evaluation score. The overall value score is as follows: 
\[
\operatorname{Value}(S, Q) =  \cdot \operatorname{Conf}(S, Q) + \cdot \operatorname{Eval}(y)
\]

\paragraph{Global Keyframe Memory.} For each node, the confidence score from node evaluation is used to decide whether keyframe descriptions should be generated. If the score exceeds a predefined threshold, \texttt{Key\_Frame\_Prompt} is used to instruct the VideoLLM to produce textual descriptions of the sampled frames.
 These descriptions are stored in a global keyframe memory to maintain cross-segment awareness and support future expansions.

\paragraph{Termination Criteria.} The search continues until a predefined budget is exhausted, such as a maximum number of iterations or when a segment’s prediction confidence exceeds a specified threshold. The final output with their corresponding responses, which can be used for downstream tasks such as video question answering. By allocating computation to high-value temporal regions, TS enables fine-grained understanding of long videos in a resource-efficient manner.



\begin{table}[ht]
\centering

\begin{tabular}{lcccc}
\toprule
\multirow{2}{*}{\textbf{Model}} & \multicolumn{2}{c}{\textbf{Video-Level}} & \multicolumn{2}{c}{\textbf{Interval-Level}} \\
\cmidrule(lr){2-3} \cmidrule(lr){4-5}
                                & \textbf{Cor.} & \textbf{Err.} & \textbf{Cor.} & \textbf{Err.} \\
\midrule
Qwen-VL-2.5    & 0.74 & 0.53 & 0.72 & 0.53 \\
LLaVA-OV      & 0.79 & 0.63 & 0.77 & 0.63 \\
LLaVA-Video    & 0.75 & 0.55 & 0.73 & 0.56 \\
\bottomrule
\end{tabular}

\caption{Correctness and error confidence scores at video-level and interval-level predictions on VideoMME.}
\label{tab:confidence_analysis}
\end{table}




\section{Experiments}


\paragraph{Experimental Setup}  
Temporal Search is configured with default settings: 5 iterations ($k = 5$), 6 expansions per step ($n = 6$), stopping thresholds $c_1 = 0.9$ and $c_2 = 0.7$, and 8 frames ($n_f = 8$) uniformly sampled per segment. We evaluate on two long-form video QA benchmarks: LongVideoBench~\cite{wu2024longvideobench} and VideoMME~\cite{fu2024video}, both featuring multi-choice questions over videos up to one hour long, posing significant challenges for efficient temporal understanding. Following the LMMEval protocol~\cite{zhang2024lmmsevalrealitycheckevaluation}, we disable subtitles during inference to isolate visual understanding and avoid reliance on textual priors. We apply TS to three representative video-based MLLMs: Qwen-VL-2.5~\cite{bai2025qwen2}, LLaVA-OV~\cite{li2024llava}, and LLaVA-Video~\cite{zhang2024videoinstructiontuningsynthetic}. All models follow a similar prompting format with a textual query, multiple answer options, and a fixed number of visual frames.

\paragraph{Baselines}  
We compare against leading video understanding models~\cite{achiam2023gpt,team2024gemini,chen2024longvila,chen2024timemarker,feng2025video} and two sampling-based baselines. Uniform Sampling (US) uniformly samples frames from the entire video for single-pass inference, following standard protocols~\cite{li2024llava,wu2024longvideobench}. Existing VideoLLMs widely adopt this strategy due to its simplicity. Uniform Temporal Voting (UTV) evenly divides the video into temporal intervals, samples a fixed number of frames from each, and performs independent inferences over each interval. Predictions with above-mean confidence are then aggregated via majority voting~\cite{wang2022self}. This serves as a static multi-inference baseline that lacks temporal awareness and iterative refinement.

\begin{figure}[t]
\centering
\includegraphics[width=1.0\columnwidth]{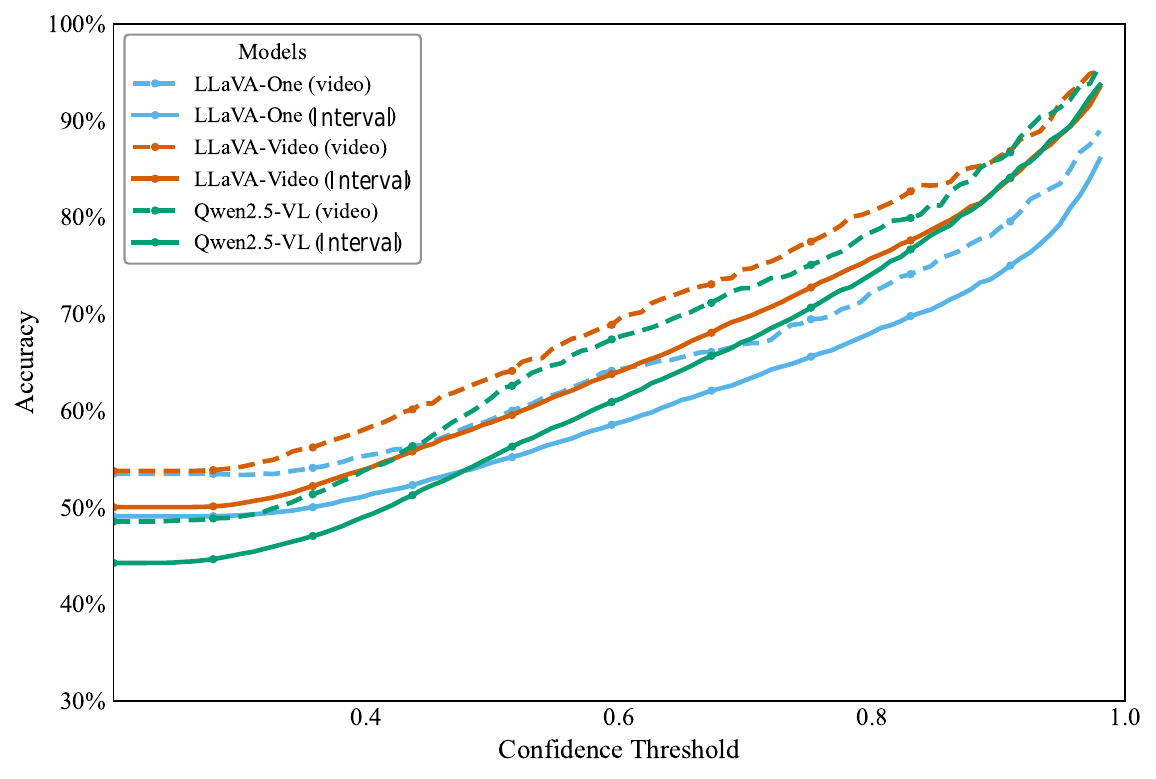} 
\caption{Relationship between model confidence and prediction accuracy at the video and interval levels. Higher confidence thresholds consistently correspond to higher accuracy, indicating that model confidence serves as a reliable signal for guiding temporal interval selection. Each point represents the accuracy of predictions whose confidence exceeds the corresponding threshold.}
\label{fig:confidence_accuracy}
\end{figure}

\subsection{Confidence–Accuracy Correlation}
We evaluate whether model-predicted confidence aligns with answer correctness under two settings, to assess its reliability as a signal for guiding temporal search. We conduct our analysis at two levels:
\begin{itemize}
    \item \textbf{Video-level:} A fixed number of frames are uniformly sampled across the entire video. The model produces a single prediction per video, and we analyze the relationship between confidence and accuracy at this global level.
    \item \textbf{Interval-level:} The video is divided into multiple temporal intervals. For each interval, the model generates an independent prediction. We then examine how confidence correlates with accuracy at the interval level.
\end{itemize}
For each prediction, we record the model's confidence and verify whether the predictions match the ground truth. Our analysis focuses on the relationship between confidence and answer accuracy.

\paragraph{Confidence–Accuracy Correlation.}
As shown in Table~\ref{tab:confidence_analysis}, all models consistently assign higher average confidence to correct predictions than to incorrect ones under both the video-level and interval-level settings. For instance, Qwen-VL-2.5 achieves an average confidence of 0.74 for correct answers, compared to 0.53 for incorrect ones under video-level sampling. Figure~\ref{fig:confidence_accuracy} further illustrates that accuracy steadily increases with higher confidence thresholds across all models and settings. These findings confirm that model confidence is a reliable indicator of prediction correctness, justifying its use in guiding temporal search for efficient long-video inference.

\begin{table}[t]
\centering
\small
\setlength{\tabcolsep}{0.5mm}  
\begin{tabular}{lccc}
\toprule
\textbf{Method} & \textbf{Frames / Inf.} & \textbf{LVB} & \textbf{V-MME} \\
\midrule

\multicolumn{4}{l}{\textit{Closed-source Models}} \\
GPT-4V~\cite{achiam2023gpt}                 & 256 & 61.3 & 59.9 \\
GPT-4o~\cite{achiam2023gpt}                 & 256 & 66.7 & 71.9 \\
Gemini-1.5-Pro~\cite{team2024gemini}        & 256 & 64.0 & 75.0 \\
Gemini-1.5-Flash~\cite{team2024gemini}      & 256 & 61.6 & 70.3 \\
\midrule
\multicolumn{4}{l}{\textit{Open-source Models}} \\
TimeMarker-8B~\cite{chen2024timemarker}     & 128 & 56.3 & 57.3 \\
LongVILA-7B~\cite{chen2024longvila}         & 256 & 57.7 & 52.1 \\
PLLaVA-34B~\cite{xu2024pllava}              &  32 & 53.2 & --   \\
Video-XL-7B~\cite{shu2024video}             & 256 & 48.8 & 51.0 \\
Video-R1-7B~\cite{feng2025video}            &  64 & --   & 52.2 \\
\midrule
Qwen2-VL-7B~\cite{wang2024qwen2}            & 256 & 55.6 & --   \\
Qwen2-VL-7B w/ TS                            &  16 & 56.6 & --   \\
\midrule
\multicolumn{4}{c}{\textbf{LLaVA-Video (7B)}} \\
\midrule
Uniform Sampling                            &   8 & 53.5 & 53.7 \\
Uniform Temporal Voting                     &   8 & 54.4 & 54.7 \\
TS                                  &   8 & 56.3 & 55.6 \\
TS-BFS                             &   8 & \textbf{57.9} & \textbf{57.5} \\
\midrule
\multicolumn{4}{c}{\textbf{LLaVA-OV (7B)}} \\
\midrule
Uniform Sampling                            &   8 & 51.6 & 53.4 \\
Uniform Temporal Voting                     &   8 & 51.6 & 53.2 \\
TS                            &   8 & 53.9& 55.4 \\
TS-BFS                             &   8 & \textbf{54.4} & \textbf{56.9} \\
\midrule
\multicolumn{4}{c}{\textbf{Qwen2.5-VL (7B)}} \\
\midrule
Uniform Sampling                            &   8 & 51.5 & 48.5 \\
Uniform Temporal Voting                     &   8 & 50.8 & 49.3 \\
TS                                  &   8 & 56.4 & 53.6 \\
TS-BFS                             &   8 & \textbf{57.9} & \textbf{55.1} \\
\bottomrule
\end{tabular}
\caption{
Comparison of different frame selection strategies on the LongVideoBench (LVB) and VideoMME (V-MME) benchmarks. “Frames / Inf.” indicates the number of frames used per inference. Our proposed \textbf{Temporal Search} consistently outperforms other baselines across different architectures, demonstrating its general applicability.
}
\label{tab:frame_selection}
\end{table}


\begin{figure}[t]
\centering
\includegraphics[width=0.45\textwidth]{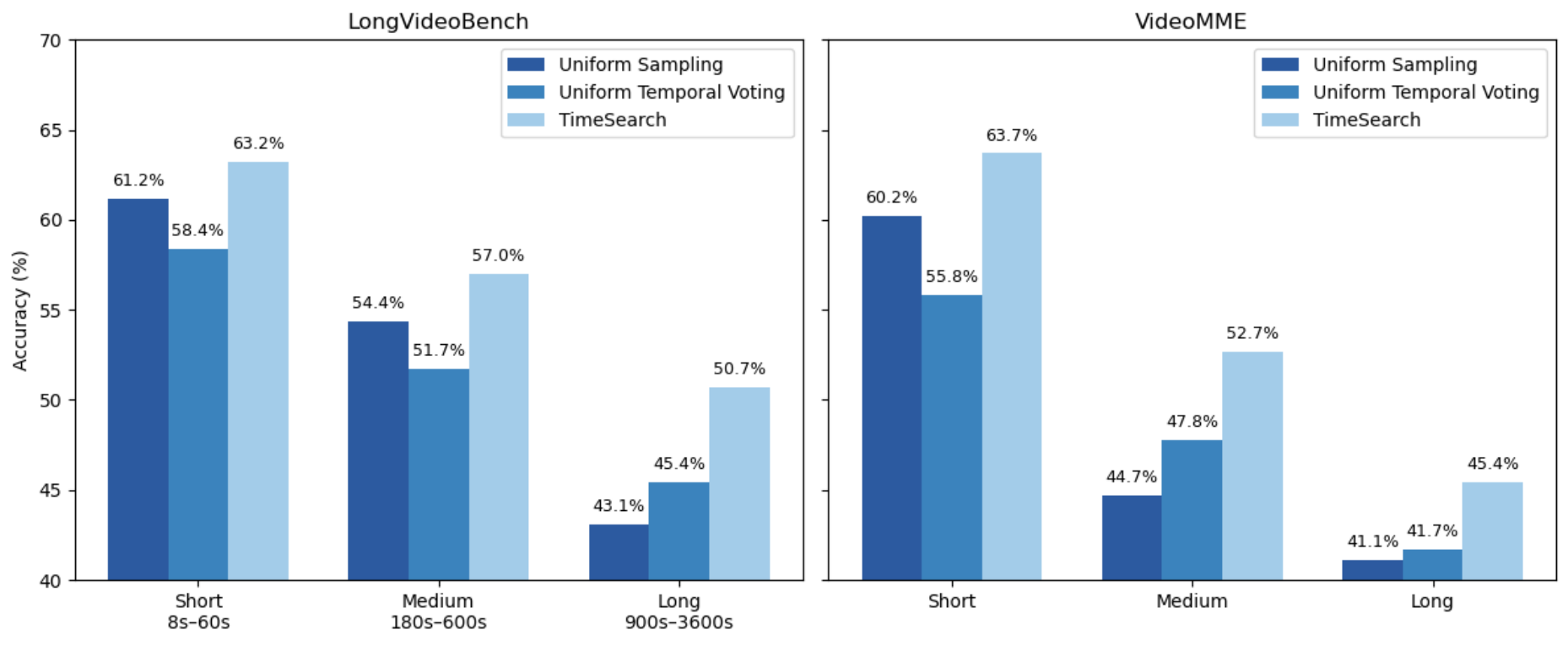} 
\caption{Accuracy on LongVideoBench and VideoMME by duration based on the Qwen2.5-Vl. TS improves accuracy across all durations, showing better generalization and robustness.}

\label{fig:duration_ablation}
\end{figure}

\subsection{Results on LongVideo Benchmark}

As shown in Table~\ref{tab:frame_selection}, TS consistently outperforms US and UTV across all models on both LongVideoBench and VideoMME. Furthermore, incorporating best-first search over a temporal interval tree further improves performance by enabling more efficient exploration. For Qwen2.5-VL, TS-BFS improves accuracy from 51.5\% to 57.9\% on LongVideoBench and from 48.5\% to 55.1\% on VideoMME. Similar gains are observed for LLaVA-Video and LLaVA-OV, confirming the general effectiveness of TS. TS achieves competitive results with only 8 frames per inference, outperforming larger models like TimeMarker-8B with 128 frames and Video-XL with 256 frames. Qwen2-VL with TS, using multiple lightweight inferences with 16 frames, also exceeds its baseline of 256 frames. In contrast, UTV yields limited improvements, highlighting the need for adaptive search where the model actively perceives and proposes relevant temporal intervals rather than relying on static aggregation. 

\begin{figure}[t]
\centering
\includegraphics[width=1.0\columnwidth]{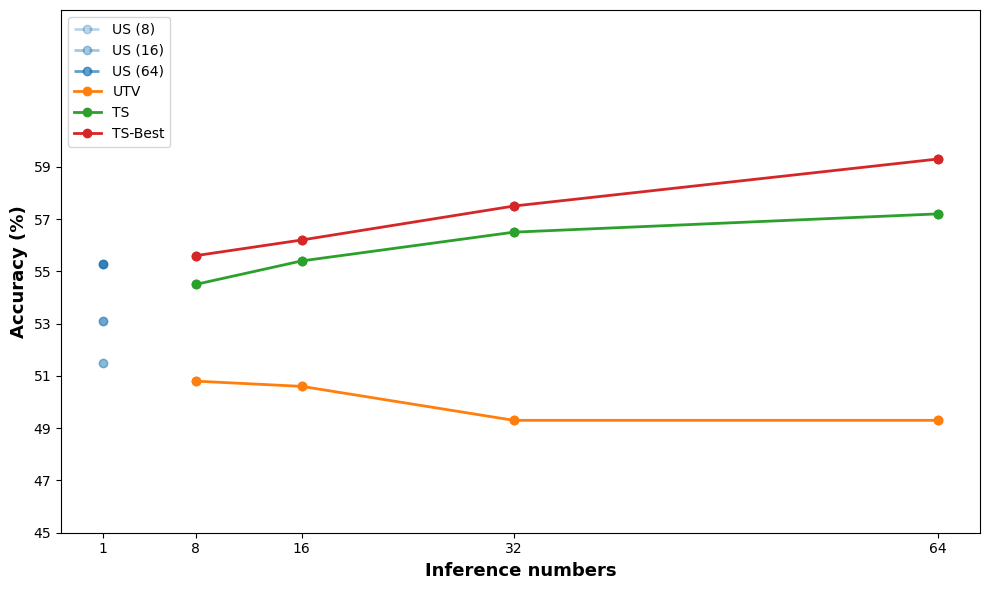} 
\caption{
Model performance under different numbers of inference passes. Results are based on the Qwen2.5-VL
on LongVideoBench.}
\label{fig:infer_num}
\end{figure}

\subsection{Inference Scaling}
We investigate how model performance scales with the number of inference passes. As shown in Figure~\ref{fig:infer_num}, Uniform Sampling (US) performs single-pass inference and shows modest improvement as the number of sampled frames increases. Although Uniform Temporal Voting (UTV) involves multiple inferences across temporal segments, it yields only marginal gains. In contrast, both TS and TS-BFS consistently improve accuracy with more inference steps, highlighting the effectiveness of confidence-guided temporal search. These results highlight the advantage of adaptive, iterative inference over static multi-pass aggregation, particularly when operating within fixed per-pass frame budgets. Moreover, the tree-based search in TS-BFS can be efficiently parallelised by evaluating multiple candidate intervals in batches. For example, expanding six nodes per step enables six parallel inference passes to be executed in a single batch.

\subsection{Ablation studies}
We conduct ablation studies to better understand the design choices and robustness of the proposed TS framework. Specifically, we examine the robustness of TS across different video durations, analyze the effect of the number of sampled frames per interval, and evaluate the influence of stopping conditions and search configurations.
\begin{table}[t]
\centering
\small
\setlength{\tabcolsep}{2.5mm}
\begin{tabular}{cccccc}
\toprule
\textbf{Frames} & \textbf{US} & \textbf{UTV} & \textbf{TS} & \textbf{TS-Best} & \textbf{Memory} \\
\midrule
8   & 51.5 & 50.8 & 56.4 & \textbf{57.9} & 17\\
16  & 53.1 & 49.2 & 57.8 & \textbf{59.8} & 22 \\
32  & 55.3 & 54.7 & 59.0 & \textbf{60.7} & 39 \\
64  & 55.3 & 55.5 & 60.1 & \textbf{61.7} & 120 \\
\bottomrule
\end{tabular}
\caption{
Accuracy and GPU memory usage (GB) of different sample methods under varying numbers of sampled frames per interval. Results are based on the Qwen2.5-VL on LongVideoBench.}
\label{tab:ablation_qwen2.5vl_frame_memory}
\end{table}

\paragraph{Impact of Video Duration.}
We evaluate the robustness of TS across videos of varying lengths, including short, medium, and long durations. As shown in Figure~\ref{fig:duration_ablation}, UTV provides slight improvements on long videos but degrades performance on short ones. In contrast, TS consistently enhances accuracy across all duration groups, demonstrating strong adaptability to different temporal scales. Specifically, on long videos, TS yields a 7.6\% accuracy improvement on LongVideoBench and a 4.3\% gain on VideoMME, highlighting its effectiveness in handling extended video.

\paragraph{Effect of Frames per Interval.}
We study how the number of sampled frames per interval affects accuracy and memory usage. As shown in Table~\ref{tab:ablation_qwen2.5vl_frame_memory}, using 8 frames may result in missing key visual cues within long intervals. Increasing the frame count to 16 provides noticeable accuracy gains, while further increases (32 or 64 frames) yield diminishing returns but significantly raise memory usage. Notably, TS-BFS with 16-frame input attains 59.8\% accuracy using just 22GB, surpassing the 55.3\% accuracy of US with 64 frames, which requires over 120GB memory. This demonstrates that TS achieves strong performance with high memory efficiency, making it practical for deployment on consumer GPUs.

\paragraph{Effect of Stopping Conditions and Search Depth.}
We examine how the stopping thresholds and search parameters affect performance. As shown in Table~\ref{tab:ablation_stopping_conditions}, moderately high values of the key-frame acceptance threshold \(c_1\)  help avoid premature convergence, while stricter early termination thresholds \(c_2\) (1.0) promote more confident final decisions. In parallel, Table~\ref{tab:ablation_kn_joint} analyses the impact of the number of search iterations \(k\) and node expansions per step \(n\). Increasing both parameters generally improves accuracy, as a deeper and broader search improves the model's ability to identify informative intervals. Performance saturates around \(k = 10\), \(n = 6\), beyond which further gains diminish relative to the increased inference cost.

\begin{table}[t]
\centering
\small
\setlength{\tabcolsep}{1.2mm}
\begin{tabular}{llcccccc}
\toprule
\multirow{2}{*}{\textbf{Model}} & \multirow{2}{*}{\(\boldsymbol{c_2}\)} & \multicolumn{6}{c}{\(\boldsymbol{c_1}\)} \\
\cmidrule(lr){3-8}
 & & \textbf{0.5} & \textbf{0.6} & \textbf{0.7} & \textbf{0.8} & \textbf{0.9} & \textbf{1.0} \\
\midrule
\multirow{6}{*}{Qwen2.5-VL} 
    & \textbf{0.5} & 53.6 & 55.0 & 55.2 & 54.7 & 53.3 & 55.6 \\
    & \textbf{0.6} & 54.5 & 55.5 & 54.8 & 55.2 & 55.4 & 56.0 \\
    & \textbf{0.7} & 52.9 & 54.5 & 56.2 & 56.2 & 55.7 & 56.6 \\
    & \textbf{0.8} & 53.5 & \textbf{55.0} & 56.6 & 56.5 & 56.7 & 57.2 \\
    & \textbf{0.9} & \textbf{53.6} & 54.8 & \textbf{57.0} & \textbf{57.5} & \textbf{57.5} & 57.4 \\
    & \textbf{1.0} & 54.4 & 55.0 & 56.6 & 57.1 & 57.4 & \textbf{58.2} \\
\midrule

\multirow{6}{*}{LLaVA-Video} 
    & \textbf{0.5} & 55.3 & \textbf{57.1} & 56.6 & 55.7 & 56.9 & 55.9 \\
    & \textbf{0.6} & 55.0 & 55.9 & 56.6 & 57.3 & 56.9 & 55.9 \\
    & \textbf{0.7} & 55.9 & 55.6 & 56.6 & 57.5 & 58.3 & 57.6 \\
    & \textbf{0.8} & 55.9 & 56.4 & \textbf{56.7} & 57.0 & 57.9 & 55.8 \\
    & \textbf{0.9} & \textbf{56.3} & 56.5 & 56.5 & 57.7 & 58.3 & 57.6 \\
    & \textbf{1.0} & 55.8 & 56.4 & 56.5 & \textbf{58.3} & \textbf{58.4} & \textbf{58.8} \\
\bottomrule
\end{tabular}
\caption{Ablation of stopping thresholds \(c_1\) and \(c_2\) for Temporal Search on LongVideoBench.}
\label{tab:ablation_stopping_conditions}
\end{table}

\begin{table}[t]
\centering
\small
\setlength{\tabcolsep}{2mm}
\begin{tabular}{llcccc}
\toprule
\textbf{Model} & \textbf{\(k \backslash n\)} & \textbf{2} & \textbf{4} & \textbf{6} & \textbf{8} \\
\midrule
\multirow{4}{*}{Qwen2.5-VL} 
    & \textbf{1}  & 53.6 & 53.3 & 55.0 & 55.0 \\
    & \textbf{5}  & 55.4 & 56.3 & 56.2 & 56.2 \\
    & \textbf{10} & \textbf{55.7} & \textbf{56.4} & \textbf{57.3} & \textbf{56.7} \\
\midrule

\multirow{4}{*}{LLaVA-Video} 
    & \textbf{1}  & 55.9 & 56.2 & 56.7 & 56.7 \\
    & \textbf{5}  & 55.5 & 56.2 & \textbf{58.1} & 57.4 \\
    & \textbf{10} & \textbf{56.5} & \textbf{56.4} & 57.8 & \textbf{57.5} \\
\bottomrule
\end{tabular}
\caption{Ablation on the number of greedy search steps \(k\) and the number of node expansions \(n\) in Temporal Search (TS). Accuracy (\%) is reported on the LongVideoBench benchmark. Each block corresponds to a different model.}
\label{tab:ablation_kn_joint}
\end{table}
\begin{figure}[t]
\centering
\includegraphics[width=0.5\textwidth]{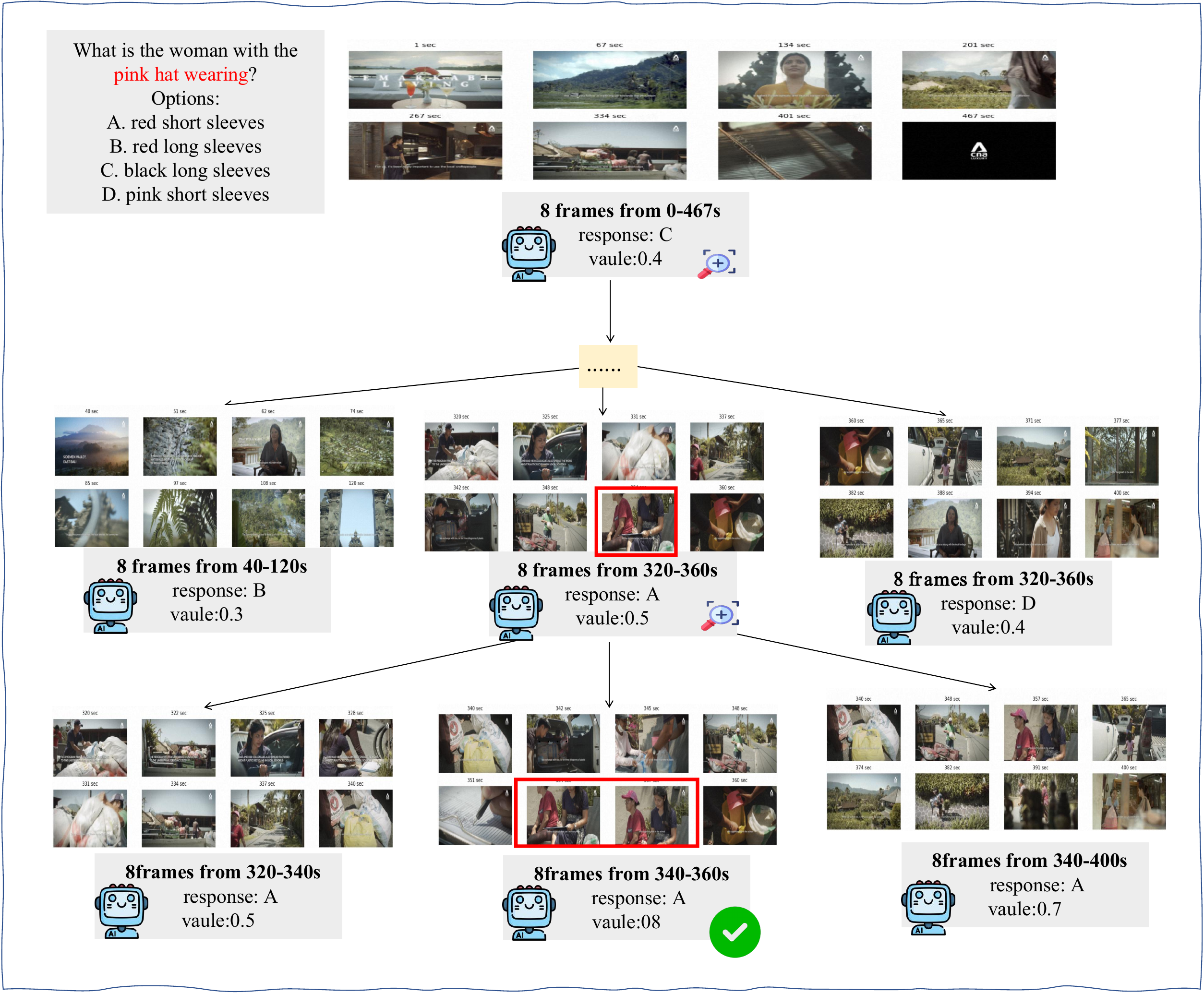} 
\caption{Illustration of the Temporal Search framework with iterative zoom-in on valuable temporal segments.}
\label{fig2}
\end{figure}

\subsection{Case Analysis}
We present a case from the LongVideoBench dataset to illustrate the behavior of our Temporal Search (TS) framework. The question is: ``What is the woman with the pink hat wearing?''
TS begins by exploring coarse-grained segments across the video. Based on model confidence, it iteratively refines its focus to more informative regions. In this example, TS converges on the time window around 350–365 seconds, where the target subject appears. These frames clearly reveal the woman in a pink hat wearing red long sleeves, allowing the model to confidently choose the correct answer: A. This example demonstrates TS's ability to efficiently zoom in on relevant content, enabling accurate reasoning in long videos.

\section{Conclusion}
In this work, we present Temporal Search (TS), a training-free framework that enables MLLMs to iteratively locate and refine task-relevant intervals in long videos. TS leverages model confidence to guide temporal exploration and focuses on finer-grained regions through fixed-frame inputs. To enhance efficiency, we introduce TS-BFS, a best-first search strategy that prioritises high-confidence intervals. TS requires no model changes and improves performance across multiple benchmarks. Experiments show that TS-BFS significantly boosts accuracy on LongVideoBench and VideoMME, demonstrating its effectiveness and generality for long-video understanding.

\bibliography{aaai2026}

\end{document}